\documentclass[runningheads]{llncs}
\usepackage[T1]{fontenc}
\usepackage{graphicx}
\usepackage{amsmath}
\usepackage{amssymb}
\usepackage{xcolor}
\usepackage{soul}
\usepackage{hyperref}
\usepackage{esvect}
\usepackage{amsmath}

\begin{document}

\title{SEDTalker: Emotion-Aware 3D Facial Animation Using Frame-Level Speech Emotion Diarization}
\titlerunning{ICPR 2026 Competition on SEDTalker}

\author{Farzaneh Jafari\inst{1}\orcidID{0009-0007-1901-2860} \and
Stefano Berretti\inst{2}\orcidID{0000-0003-1219-4386} \and
Anup Basu\inst{1}\orcidID{0000-0002-7695-4148}}

\authorrunning{F. Jafari et al.}

\institute{University of Alberta, Edmonton, AB, Canada\\
\email{farzane1@ualberta.ca}\\
\email{basu@ualberta.ca}\\
\url{https://mrc.science.ualberta.ca/} \and
University of Florence, Florence, Italy\\
\email{stefano.berretti@unifi.it}}

\maketitle  

\begin{figure*}[h]
\centering
\includegraphics[width=\textwidth]{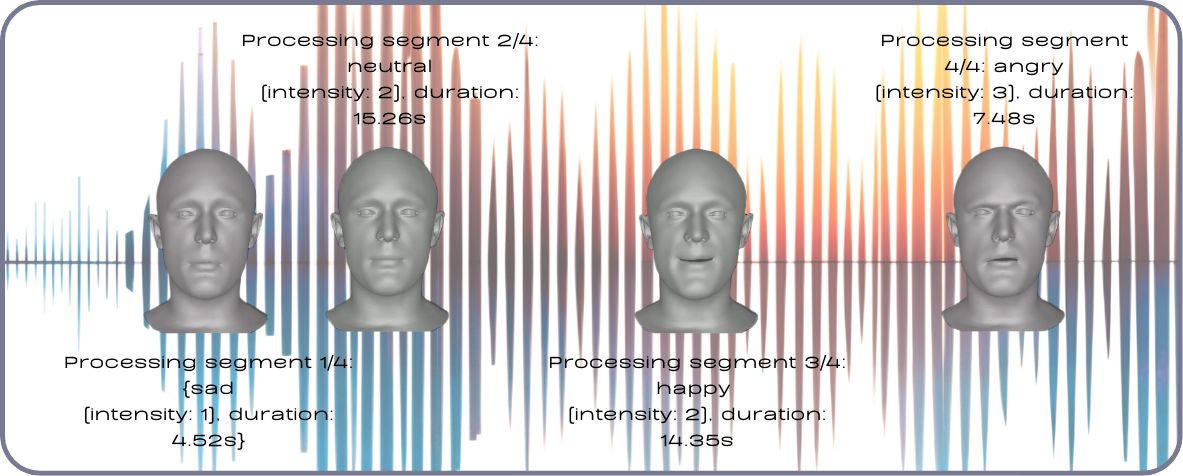}
\caption{\textbf{SEDTalker} is a speech-driven framework for emotion-aware 3D facial animation. By leveraging frame-level speech emotion diarization, the model captures fine-grained emotional cues and directly translates them into 3D facial parameters for precise expressive control.}
\label{fig:SEDTalker}
\end{figure*}

\begin{abstract}
We introduce SEDTalker, an emotion-aware framework for speech-driven 3D facial animation that leverages frame-level speech emotion diarization to achieve fine-grained expressive control. Unlike prior approaches that rely on utterance-level or manually specified emotion labels, our method predicts temporally dense emotion categories and intensities directly from speech, enabling continuous modulation of facial expressions over time. The diarized emotion signals are encoded as learned embeddings and used to condition a speech-driven 3D animation model based on a hybrid Transformer-Mamba architecture. This design allows effective disentanglement of linguistic content and emotional style while preserving identity and temporal coherence. We evaluate our approach on a large-scale multi-corpus dataset for speech emotion diarization and on the EmoVOCA~\cite{nocentini2025emovoca} dataset for emotional 3D facial animation. Quantitative results demonstrate strong frame-level emotion recognition performance and low geometric and temporal reconstruction errors, while qualitative results show smooth emotion transitions and consistent expression control. These findings highlight the effectiveness of frame-level emotion diarization for expressive and controllable 3D talking head generation. Our code and videos are available at \href{https://farzanehjafari1987.github.io/SEDTalker.github.io/}{https://SEDTalker.github.io/}.

\end{abstract}

\section{Introduction}
Speech-driven 3D facial animation has made significant progress in recent years, enabling realistic lip synchronization and identity-preserving motion from raw audio signals. Modern applications span virtual assistants, digital avatars, film production, and telepresence systems, where generating emotionally expressive and temporally coherent facial animations is essential for natural human-computer interaction. However, while existing methods excel at speech-to-motion mapping, they struggle to capture the dynamic and nuanced nature of emotional expression in speech.

A fundamental challenge lies in the temporal granularity of emotion modeling. Human emotional expression is inherently dynamic; emotions shift, blend, and vary in intensity throughout an utterance. Yet most existing methods condition animation models on a single global emotion category for the entire utterance, limiting expressiveness and failing to capture natural emotion transitions. This coarse-grained approach results in facial animations that appear emotionally flat or exhibit abrupt, unnatural expression changes. Furthermore, the scarcity of high-quality emotional speech datasets exacerbates this problem. Datasets like EmoVOCA~\cite{nocentini2025emovoca} provide richly expressive facial motion but only neutral speech, preventing models from learning direct correlations between emotional acoustic features and facial expressions. This data limitation has been a persistent bottleneck in developing truly emotion-aware talking head systems.

In this work, we address these limitations by introducing Speech Emotion Diarization (SED) as a core component for emotional facial animation. Unlike utterance-level classification, SED predicts emotion categories and intensities at fine temporal resolution directly from speech, providing dense, time-varying affective signals that capture emotion transitions, intensity fluctuations, and mixed emotional states within utterances, capabilities impossible with coarse labels. To overcome data scarcity, we propose a decoupled training strategy: the animation model trains on neutral speech paired with emotional facial sequences, while a pre-trained SED model extracts emotion signals at inference time. This approach enables effective emotion transfer without requiring aligned emotional speech-face training pairs and makes the framework adaptable to new emotions without retraining.

The main contributions of this work are the following: 1) We formulate speech-driven 3D facial animation as a frame-level emotion-conditioned generation task, enabling continuous, intra-utterance emotion modulation that utterance-level approaches cannot capture by design. 2) By separating SED from animation training, we eliminate the requirement for paired emotional speech-3D mesh data, a bottleneck that has blocked prior work. 3) We construct and release a frame-level emotion diarization corpus of 58,834 utterances aggregated from nine publicly available datasets, covering 149 speakers, four English accents, and diverse recording conditions, achieving a 78.9\% weighted F1 at 20\,ms resolution. 4) We extend the JambaTalk~\cite{jafari2024jambatalk} backbone with an additive emotion–intensity conditioning module that separates linguistic content from emotional style while maintaining temporal coherence and reduces GPU memory usage to 13\,GB for 18 emotion–intensity states plus a neutral state.

\section{Related Work}
\noindent\textbf{Speech-Driven 3D Facial Animation.}
Recent deep learning approaches have achieved impressive results in generating realistic talking heads from speech. Early works such as VOCA \cite{nocentini2025emovoca} introduced neural networks for voice-operated character animation, laying the foundation for learning-based audio-to-mesh synthesis. Transformer-based architectures have since dominated the field: FaceFormer \cite{fan2022faceformer} applies cross-modal attention between audio and motion sequences to capture long-range dependencies; CodeTalker \cite{xing2023codetalker} introduces discrete motion priors through VQ-VAE to improve motion quality; SelfTalk \cite{peng2023selftalk} proposes self-supervised commutative training for better disentanglement of content and style. More recently, diffusion-based approaches like FaceDiffuser \cite{stan2023facediffuser} leverage denoising probabilistic models for diverse motion generation. While these methods achieve high-fidelity lip synchronization and identity preservation, they typically operate on neutral or weakly emotional speech and lack explicit mechanisms for fine-grained emotional control.

\noindent\textbf{Emotion-Aware Facial Animation.}
Several works have attempted to incorporate emotion into speech-driven animation. EmoTalk \cite{peng2023emotalk} introduces emotional disentanglement by conditioning on utterance-level emotion labels and style vectors, achieving better emotional expressiveness compared to neutral baselines. However, EmoTalk \cite{peng2023emotalk} relies on a single emotion label per utterance and requires manually annotated or pre-classified emotion categories at training time. These approaches use global emotion embeddings or style codes, which cannot capture intra-utterance emotion dynamics. More critically, these approaches require paired emotional speech and facial motion data, which severely limits their applicability given dataset scarcity. In contrast, our method predicts frame-level emotion signals directly from speech at inference time and supports continuous emotion intensity control without requiring emotional speech during training.

\noindent\textbf{Speech Emotion Recognition (SER).}
Traditional SER systems classify entire utterances into discrete emotion categories. Recent advances leverage self-supervised speech representations: Wav2Vec 2.0 \cite{baevski2020wav2vec2} and WavLM \cite{chen2022wavlm} pre-trained on large-scale unlabeled speech provide robust acoustic features that transfer well to emotion recognition tasks. However, most SER works operate at utterance-level granularity, producing a single emotion label per audio segment. Frame-level emotion recognition, predicting emotion at fine temporal resolution, has received less attention despite its importance for capturing temporal dynamics. Our work bridges this gap by training a frame-level emotion diarization model on a multi-corpus dataset of 58,834 utterances and demonstrating its effectiveness for downstream animation conditioning.

\begin{table}[t]
\setlength{\tabcolsep}{8pt}
\centering
\caption{Dataset composition and characteristics.}
\label{tab:datasets}
\begin{tabular}{lcccc}
\hline
\textbf{Dataset} & \textbf{Utterances} & \textbf{Style} & \textbf{Speakers} & \textbf{Contrib.\%} \\
\hline
MELD~\cite{poria2019meld} & 12,070 & Conv. & Multi & 20.5 \\
IEMOCAP~\cite{busso2008iemocap} & 11,032 & Script+Improv & 10 & 18.7 \\
JL-Corpus~\cite{james2014jl} & 10,661 & Acted & 4 & 18.1 \\
ESD~\cite{zhou2021esd} & 10,500 & Professional & 10 & 17.8 \\
CREMA-D~\cite{cao2014crema} & 7,438 & Acted & 91 & 12.6 \\
EmoV-DB~\cite{adigwe2018emovdb} & 3,353 & Varied & 4 & 5.7 \\
TESS~\cite{dupuis2010tess} & 2,400 & Acted & 2 & 4.1 \\
RAVDESS~\cite{livingstone2018ravdess} & 960 & Acted & 24 & 1.6 \\
SAVEE~\cite{haq2009savee} & 420 & Acted & 4 & 0.7 \\
\hline
\textbf{Total} & \textbf{58,834} & \textbf{Diverse} & \textbf{149} & \textbf{100} \\
\hline
\end{tabular}
\end{table}

\begin{table}[t]
\setlength{\tabcolsep}{8pt}
\centering
\caption{Emotion distribution and normalized class weights after frame-level propagation (20\,ms).}
\label{tab:distribution_frames}
\begin{tabular}{lccccc}
\hline
\textbf{Emotion} & \textbf{Utt.} & \textbf{Utt.\%} & \textbf{Frames (M)} & \textbf{Frame\%} & \textbf{Weight} \\
\hline
Happy    & 16,133 & 27.4 & 4.00 & 32.1 & 0.31 \\
Angry    & 15,115 & 25.7 & 4.05 & 32.4 & 0.34 \\
Sad      & 10,558 & 17.9 & 2.30 & 18.5 & 0.48 \\
Neutral  &  8,041 & 13.7 & 0.20 &  1.6 & 0.63 \\
Upset    &  3,573 &  6.1 & 1.12 &  9.0 & 1.42 \\
Disgust  &  3,054 &  5.2 & 0.53 &  4.3 & 1.66 \\
Fear     &  2,360 &  4.0 & 0.55 &  2.0 & 2.15 \\
\hline
\textbf{Total} & \textbf{58,834} & \textbf{100} & \textbf{12.45} & \textbf{100} & \textbf{7.0} \\
\hline
\end{tabular}
\end{table}

\section{Speech Emotion Diarization}
The frame-level emotion recognition system is based on the SpeechBrain \cite{ravanelli2021speechbrain} framework, which consists of three main components: a pre-trained WavLM feature extractor, a transformer encoder with selective fine-tuning, and a frame-level classification head.

\noindent\textbf{Feature Extraction.}
We employ WavLM-Base-Plus \cite{chen2022wavlm} as our acoustic feature extractor. WavLM is a self-supervised speech representation model pre-trained on 94,000 hours of unlabeled speech data using masked prediction objectives. The model processes raw 16\,kHz audio waveforms through a 7-layer convolutional neural network feature encoder, followed by layer normalization and a GELU activation layer, producing 512-dimensional latent features at a 20\.ms stride. These features are then passed through a 12-layer transformer encoder with 768 hidden dimensions, 8 attention heads, and 3072 feed-forward dimensions per layer.

\noindent\textbf{Selective Fine-Tuning.}
To balance feature adaptation and computational efficiency, we employ selective fine-tuning. The first 6 transformer layers remain frozen with pre-trained weights, preserving general-purpose speech representations. The upper 6 Transformer layers are fine-tuned on emotion-specific features, allowing adaptation to the emotion recognition task. The convolutional feature extractor remains frozen to preserve low-level acoustic feature extraction. This strategy reduces trainable parameters from 95M to 12.4M (13\% of total).

\noindent\textbf{Classification Head.}
Frame-level emotion classification is performed via a multi-layer perceptron:
\begin{gather}
\mathbf{h}^{(1)} = \text{ReLU}(\mathbf{W}_1 \mathbf{f} + \mathbf{b}_1) \\
\mathbf{h}^{(2)} = \text{Dropout}(\mathbf{h}^{(1)}, p=0.3) \\
\mathbf{h}^{(3)} = \text{ReLU}(\mathbf{W}_2 \mathbf{h}^{(2)} + \mathbf{b}_2) \\
\mathbf{h}^{(4)} = \text{Dropout}(\mathbf{h}^{(3)}, p=0.3) \\
\mathbf{z} = \mathbf{W}_3 \mathbf{h}^{(4)} + \mathbf{b}_3 \\
\mathbf{p} = \text{Softmax}(\mathbf{z})
\end{gather}
\noindent where $\mathbf{f} \in \mathbb{R}^{768}$ is the WavLM feature vector, $\mathbf{W}_1 \in \mathbb{R}^{512 \times 768}$, $\mathbf{W}_2 \in \mathbb{R}^{128 \times 512}$, $\mathbf{W}_3 \in \mathbb{R}^{7 \times 128}$ are learnable weight matrices, and $\mathbf{p} \in \mathbb{R}^7$ is the predicted probability distribution. Layer normalization and dropout ($p=0.3$) provide regularization.

\subsection{Data Preparation}
We construct a large-scale emotion corpus by aggregating nine publicly available English speech datasets, ensuring robust cross-corpus generalization. The corpus comprises 58,834 utterances spanning diverse recording conditions, speaker demographics, and emotion elicitation methods (acted, scripted, improvised, conversational). The corpus covers four English accents (US, UK, Canadian, New Zealand), age range 20--70 years, and balanced gender representation (48\% female, 52\% male), providing comprehensive diversity for robust generalization. Table~\ref{tab:datasets} summarizes the composition.

\noindent\textbf{Emotion Mapping.}
We adopt a seven-class discrete emotion taxonomy, \(\mathcal{E} = \{\textit{angry, disgust, fear, happy, neutral, sad, upset}\}\), based on Ekman's basic emotions~\cite{ekman1992basic}, augmented with an \textit{upset} category to capture frustration-like affective states that are not well represented by the canonical classes. The \textit{upset} class corresponds to the \textit{frustrated} label in the IEMOCAP dataset~\cite{busso2008iemocap}, while a \textit{neutral} baseline is included to model emotionally unmarked speech. Dataset-specific emotion labels are systematically mapped to this unified taxonomy (e.g., \textit{frustrated} $\rightarrow$ \textit{upset}, \{\textit{excited, amused}\} $\rightarrow$ \textit{happy}, \textit{calm} $\rightarrow$ \textit{neutral}).

\noindent\textbf{Class Imbalance and Weighting.}
The corpus exhibits severe class imbalance (Table~\ref{tab:distribution_frames}), with \textit{happy} (27.4\%) dominating and \textit{fear} (4.0\%) underrepresented at the utterance level. This imbalance intensifies at the frame level due to duration bias: high-arousal emotions (\textit{angry}, \textit{happy}, \textit{upset}) average 6-7 seconds, while \textit{neutral} averages only 2.1 seconds. We employ inverse frequency weighting to mitigate class imbalance. 
For each emotion class $c$, the weight is computed as follows:
\begin{equation}
w_c = \frac{N_{\text{total}}}{K \cdot N_c},
\label{eq:weight}
\end{equation}
\noindent where $N_{\text{total}}=41{,}181$ is the total number of training samples, $K=7$ is the number of emotion classes, and $N_c$ is the number of samples for class $c$. It computes inverse frequency weighting and normalizes weights to sum to $K$. This ensures minority classes (\textit{fear}: $w=2.15$) receive higher loss contributions than majority classes (\textit{happy}: $w=0.31$). The weighted frame-level cross-entropy loss is:
\begin{equation}
\mathcal{L} = -\frac{1}{F}\sum_{f=1}^{F} w_{c_f} \log p(e_f = c_f \mid \mathbf{x}_f),
\label{eq:loss}
\end{equation}
\noindent where $F$ is the total number of frames in a batch, $c_f$ is the ground truth emotion for frame $f$, $w_{c_f}$ is the weight for class $c_f$, and $p(e_f = c_f \mid \mathbf{x}_f)$ is the model's predicted probability that frame $f$ has emotion $c_f$ given audio features $\mathbf{x}_f$. This weighted loss amplifies gradients for rare emotions during backpropagation, forcing the model to learn discriminative features for all classes, despite severe imbalance.

\noindent\textbf{Frame-Level Propagation.}
We perform stratified splitting of the data into 70\% training, 15\% validation, and 15\% test sets, preserving the exact emotion distribution across splits (Table~\ref{tab:distribution_frames}). Each split retains representation from all nine source datasets, preventing dataset-specific overfitting. For frame-level training at 20\,ms resolution, utterance labels are propagated uniformly: given an utterance of duration $t$ seconds with emotion $e$, we generate $\lfloor t/0.02 \rfloor$ frame labels, all set to $e$.

\subsection{Training Configuration}
The model is trained using weighted cross-entropy loss with the AdamW optimizer (learning rate $1 \times 10^{-4}$, weight decay $1 \times 10^{-4}$). The learning rate is adaptively reduced on validation plateau using ReduceLROnPlateau (factor=0.5, patience=3). To increase the effective batch size, gradients are accumulated over 8 steps with a physical batch size of 4, yielding an effective batch size of 32. Mixed precision training is employed to accelerate computation, and regularization is applied via dropout ($p=0.3$). Training converges at epoch 50, achieving a test accuracy of 78.92\% after approximately 30 hours on an NVIDIA RTX 4090 (24GB VRAM).

\section{Methodology}
\subsection{Problem Formulation}
We formulate the problem of speech-driven 3D emotional facial animation as learning a mapping function $\mathcal{F}: (\mathbf{A}, \mathbf{T}, e, i) \rightarrow \mathbf{V}$ that generates a sequence of 3D facial vertices $\mathbf{V} \in \mathbb{R}^{T \times N \times 3}$ from input audio $\mathbf{A}$, a subject-specific template mesh $\mathbf{T} \in \mathbb{R}^{N \times 3}$, emotion label $e \in \{0, 1, \ldots, E-1\}$, and an intensity level $i \in \mathbb{R}^+$, where $T$ is the sequence length, $N=5,023$ is the number of vertices in the FLAME topology, and $E$ is the number of emotion categories.

Given an audio waveform sampled at 16 kHz, our objective is to synthesize temporally coherent 3D facial meshes at 30 FPS that exhibit both speech-synchronized lip movements and emotion-consistent expressions. The model must learn to disentangle speech content from emotional style while maintaining subject identity through the template mesh. We optimize the following composite loss function:
\begin{equation}
\mathcal{L} = \lambda_1 \mathcal{L}_{\text{vert}} + \lambda_2 \mathcal{L}_{\text{vel}} + \lambda_3 \mathcal{L}_{\text{lip}} + \lambda_4 \mathcal{L}_{\text{ctc}},
\end{equation}
\noindent where $\mathcal{L}_{\text{vert}}$ ensures vertex-level reconstruction accuracy, $\mathcal{L}_{\text{vel}}$ enforces temporal smoothness through velocity constraints, $\mathcal{L}_{\text{lip}}$ aligns lip movements with phonetic features, and $\mathcal{L}_{\text{ctc}}$ provides text-level supervision. The loss weights are set to $\lambda_1=1000$, $\lambda_2=1000$, $\lambda_3=0.001$, and $\lambda_4=0.0001$ based on empirical validation.

\subsection{Architecture}
The architecture is composed of four core modules: audio feature extraction, emotion conditioning module, hybrid sequential modeling, and vertex reconstruction.

\noindent\textbf{Audio Encoder.}
We adopt the pre-trained Wav2Vec 2.0 \cite{baevski2020wav2vec2} model as our audio encoder, which extracts robust acoustic representations from raw speech waveforms. The encoder maps input audio to 1024-dimensional frame-level features at approximately 50 FPS, which are temporally resampled to 30 FPS using linear interpolation to match the target animation frame rate. To preserve speech content information, we additionally employ a parallel frozen Wav2Vec2-CTC model that produces phonetic alignments and text-level features for lip synchronization supervision.

\noindent\textbf{Emotion Conditioning Module.}
Emotion conditioning is achieved through learned embeddings that modulate the audio features before sequential processing. We represent discrete emotion categories through an embedding layer $\mathbf{E}_e \in \mathbb{R}^{E \times d}$ where $d=512$ is the feature dimension, and continuous intensity values through a learned projection $\mathbf{W}_i \in \mathbb{R}^{1 \times d}$. The conditioning vector is computed as:
\begin{equation}
\mathbf{C} = \text{Emb}(e) + \mathbf{W}_i \cdot i,
\end{equation}
\noindent where $\text{Emb}(e)$ retrieves the emotion-specific embedding. This conditioning vector is broadcast across the temporal dimension and added to the projected audio features:
\begin{equation}
\mathbf{H}_{\text{cond}} = \mathbf{W}_{\text{proj}}\mathbf{H}_{\text{audio}} + \mathbf{C},
\end{equation}
\noindent where $\mathbf{W}_{\text{proj}} \in \mathbb{R}^{1024 \times 512}$ projects audio features to the model's hidden dimension. This additive conditioning strategy allows the model to modulate speech-driven dynamics based on emotional context without disrupting the learned speech-to-motion mappings.

\noindent\textbf{Hybrid Sequential Backbone.}
We build upon the hybrid Transformer–Mamba architecture of JambaTalk \cite{jafari2024jambatalk}, which combines Mamba layers for efficient linear-complexity sequential modeling ($\mathcal{O}(T)$) with Transformer layers. The backbone processes emotion-conditioned features $\mathbf{H}_{\text{cond}}$ through an initial MoE-Mamba layer, a Transformer layer, and a final MoE-Mamba layer. Each MoE-Mamba layer employs two experts with top-2 routing, enabling specialized modeling of different facial regions \cite{jafari2024jambatalk}.

\noindent\textbf{Vertex Reconstruction and Lip Refinement.}
The final hidden states are projected to vertex displacements through a zero-initialized linear layer, where the predicted vertices are computed by adding the transformed hidden representations to the subject-specific template mesh. To enhance lip synchronization, we employ a specialized lip refinement pathway that extracts the 254 FLAME mouth vertices and processes them through a Transformer to produce lip-specific features. These features are upsampled to 50 FPS and supervised against the text encoder features via MSE loss. Additionally, a CTC head predicts character sequences from the lip features for explicit phonetic alignment, computing the CTC loss against target character sequences. 

The complete model contains approximately 457M trainable parameters and is optimized end-to-end using Adam (learning rate $10^{-4}$) with gradient accumulation (batch size 1) for 50 epochs on EmoVOCA \cite{nocentini2025emovoca}.

\subsection{Inference Phase}
During inference, an emotional speech signal $x(t)$ is first processed by a Speech Emotion Diarization (SED) model with intensity estimation, producing a temporally ordered set of emotion segments $\mathcal{S} = \{(s_i, e_i, c_i, \ell_i)\}_{i=1}^{N}$, where $s_i$ and $e_i$ denote segment boundaries, $c_i \in \mathcal{C}$ is the emotion class, and $\ell_i \in \{1, 2, 3\}$ is the corresponding intensity level. To enforce temporal consistency, the raw SED outputs are smoothed by removing short segments, merging adjacent segments with identical $(c_i, \ell_i)$ pairs, and filling gaps to form a continuous emotion timeline. The full audio waveform is encoded into frame-level speech representations $\mathbf{A} = \{a_t\}_{t=1}^{T}$ using a pre-trained Wav2Vec 2.0 \cite{baevski2020wav2vec2} encoder. For each segment, an emotion–intensity embedding $e_{c_i,\ell_i}$ is generated and linearly interpolated at segment transitions, yielding a conditioning sequence $\mathbf{C} = \{c_t\}_{t=1}^{T}$, where $c_t = (1 - \alpha_t)e_{c_{i-1},\ell_{i-1}} + \alpha_t e_{c_i,\ell_i}$ within transition regions. The conditioned features $\tilde{\mathbf{A}} = \mathbf{A} + \mathbf{C}$ are processed by the JambaTalk \cite{jafari2024jambatalk} sequence backbone to predict frame-wise 3D vertex displacements $\Delta \mathbf{V} = \{\Delta v_t\}_{t=1}^{T}$. Finally, the predicted offsets are added to a subject-specific neutral template mesh $\mathbf{V}_0$, producing $\mathbf{V}_t = \mathbf{V}_0 + \Delta v_t$, which can be rendered as temporally coherent and emotion-aware and synchronized with the input speech.

\section{Experiments}
\subsection{Speech Emotion Diarization Evaluation}
We evaluate the trained SED model on the held-out test set comprising 7,007 utterances (1.85M frames, 10.3 hours). All metrics are computed at 20\,ms frame resolution with temporal smoothing (window size = 5 frames, corresponding to 100\,ms). The model achieves a frame-level accuracy of 78.92\% and a weighted F1-score of 78.9\%, indicating strong overall performance across emotion categories. Table~\ref{tab:frame_results} reports detailed per-emotion precision, recall, and F1-scores.

\begin{table}[t]
\setlength{\tabcolsep}{8pt}
\centering
\caption{Frame-level emotion recognition results on the test set.}
\label{tab:frame_results}
\begin{tabular}{lccccc}
\hline
\textbf{Emotion} & \textbf{Precision} & \textbf{Recall} & \textbf{F1} & \textbf{Support} & \textbf{Frames \%} \\
\hline
Happy   & 0.822 & 0.773 & 0.797 & 595,723 & 32.2 \\
Angry   & 0.797 & 0.824 & 0.810 & 594,488 & 32.1 \\
Sad     & 0.766 & 0.736 & 0.750 & 351,115 & 19.0 \\
Neutral & 0.766 & 0.879 & 0.819 & 24,052  & 1.3 \\
Upset   & 0.694 & 0.828 & 0.755 & 169,368 & 9.2 \\
Disgust & 0.837 & 0.915 & 0.874 & 77,918  & 4.2 \\
Fear    & 0.824 & 0.496 & 0.620 & 36,999  & 2.0 \\
\hline
\textbf{Weighted} & \textbf{0.793} & \textbf{0.789} & \textbf{0.789} & \textbf{1,849,663} & \textbf{100} \\
\hline
\end{tabular}
\end{table}

\begin{figure}[t]
\centering
\includegraphics[width=\columnwidth]{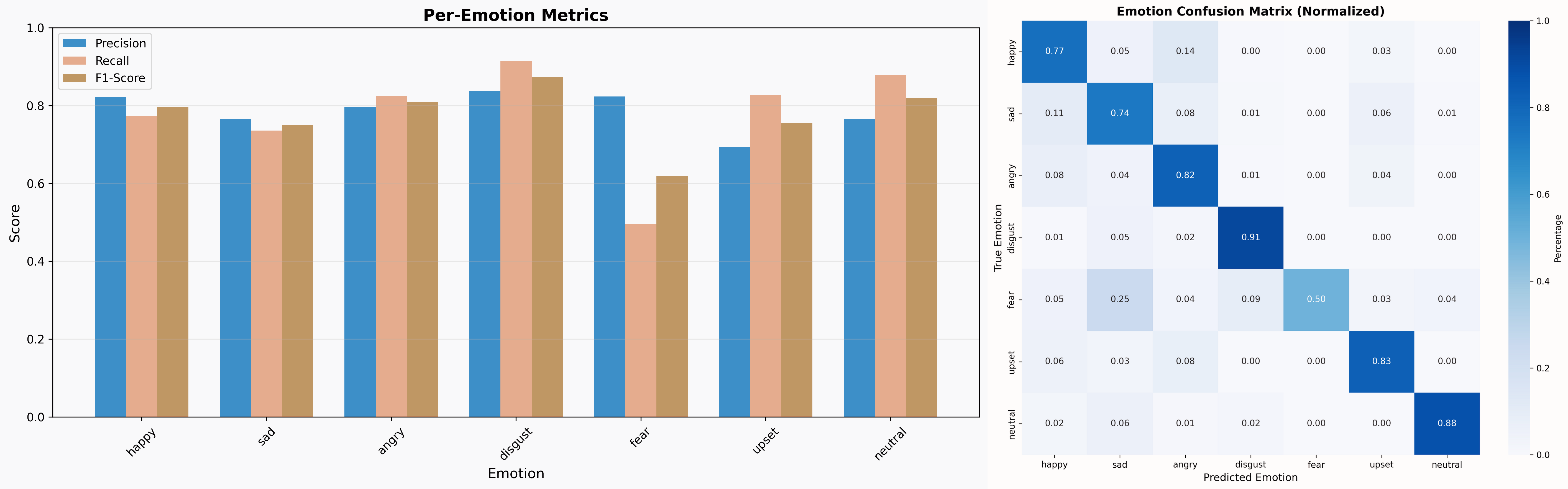}
\caption{\textit{Left:} Per-emotion precision, recall, and F1-scores on the test set. \textit{Disgust} achieves the highest F1 (87.4\%), while \textit{fear} remains the most challenging class (62.0\%) due to data scarcity. Error bars denote 95\% confidence intervals estimated via 1,000 bootstrap resamples. 
\textit{Right:} Normalized confusion matrix on the test set. Values represent row-wise percentages (true emotion $\rightarrow$ predicted emotion). Diagonal entries indicate correct classifications. Major confusions occur between high-arousal emotions (\textit{happy} $\leftrightarrow$ \textit{angry}) and between \textit{fear} and \textit{sad}.}
\label{fig:confusion_matrix}
\end{figure}

Figure~\ref{fig:confusion_matrix} (\textit{Left}) visualizes the per-emotion precision, recall, and F1-scores. 
\textit{Happy} achieves an F1-score of 79.7\%, with high precision (82.2\%) but moderately lower recall (77.3\%), indicating conservative predictions that reduce false positives. 
\textit{Angry} performs strongly with an F1-score of 81.0\%, benefiting from its distinctive high-energy and fast-tempo acoustic characteristics. 
\textit{Sad} reaches an F1-score of 75.0\%, showing balanced precision–recall behavior despite acoustic overlap with other low-arousal emotions. 
\textit{Neutral} attains an F1-score of 81.9\%, with particularly high recall (87.9\%), reflecting its stable prosodic structure and low variability. 
\textit{Upset} achieves an F1-score of 75.5\%, maintaining a reasonable balance between precision and recall despite semantic and acoustic overlap with both \textit{sad} and \textit{angry} categories. 
\textit{Disgust} yields the highest performance with an F1-score of 87.4\% and recall of 91.5\%, demonstrating that the applied class weighting effectively compensates for its low representation (4.2\% of test frames). 
\textit{Fear} remains the most challenging class, achieving an F1-score of 62.0\%. Although precision is high (82.4\%), recall is limited (49.6\%), indicating that nearly half of \textit{fear} frames are misclassified. This limitation primarily stems from severe data scarcity (2.0\% of frames) and acoustic similarity to other low-arousal emotions.

Figure~\ref{fig:confusion_matrix} (\textit{Right}) presents the normalized confusion matrix. \textit{Happy} and \textit{angry}, both high-arousal emotions with opposing valence, exhibit the most prominent bidirectional confusion (14\% \textit{happy} $\rightarrow$ \textit{angry} and 8\% \textit{angry} $\rightarrow$ \textit{happy}). This indicates that the model tends to prioritize arousal-related cues over valence in ambiguous cases. \textit{Fear} is most frequently confused with \textit{sad} (25\%), followed by \textit{disgust} (9\%) and \textit{happy} (5\%). The confusion between \textit{fear} and \textit{sad} reflects shared low-energy and subdued prosodic patterns, while confusion with \textit{disgust} arises from similar negative vocal characteristics. \textit{Sad} also shows moderate confusion with \textit{upset} (6\%) and \textit{happy} (11\%), suggesting difficulty in distinguishing emotions with overlapping arousal levels when valence cues are subtle. Overall, the confusion matrix exhibits strong diagonal dominance (50--91\% per class), confirming robust discrimination performance with errors concentrated among semantically and acoustically related emotion pairs.

\subsection{Emotional Face Animation Dataset}
We conduct our experiments using the EmoVOCA~\cite{nocentini2025emovoca} dataset, an emotion-enriched extension of Vocaset \cite{cudeiro2019vocaset} for speech-driven 3D facial animation. It provides synchronized audio, emotion annotations, and high-quality 3D facial mesh sequences with consistent topology across multiple affective states. However, a critical limitation is that the input audio remains neutral, creating a mismatch between emotional facial expressions and speech prosody. This constraint motivates our decoupled training strategy, where the animation model learns from neutral speech paired with emotional faces, while emotion signals are extracted separately at inference time.

We exclude several non-standard labels from EmoVOCA~\cite{nocentini2025emovoca} that do not align with established emotion taxonomies \cite{ekman1992basic}, including physiological states (e.g., "Drunk2", "Ill"), general moods (e.g., "Moody"), visual expressions without speech correlates (e.g., "Smile2"), and context-dependent states (e.g., "Suspicious"). Our experiments focus on the standard emotion categories that have clear acoustic manifestations in speech.

\subsection{Quantitative Evaluation}
Our model consists of 457,041,278 parameters and is evaluated on a total of 5,544 sequences comprising 628,668 frames. The overall performance in Table~\ref{tab:metrics_by_emotion_scaled} indicates that the proposed approach generates highly accurate and temporally stable facial motion. Specifically, we obtain a Mean Vertex Error (MVE) of $5.976\times10^{-3}$, demonstrating that the reconstructed meshes remain close to the ground-truth geometry. Region-specific metrics further confirm this accuracy, with a Lip Vertex Error (LVE) of $7.847\times10^{-3}$ and an Emotion-related Vertex Error (EVE) of $7.863\times10^{-3}$, indicating reliable modeling of both articulatory and expressive facial regions. Temporal metrics show smooth and coherent motion trajectories, reflected by a Motion Offset Deviation (MOD) of $1.586\times10^{-4}$, an Acceleration Error (AE) of $1.576\times10^{-4}$, and a Temporal Consistency (TC) score of $1.209\times10^{-4}$. The Fourier Frequency Error (FFE) $1.848\times10^{-3}$ further validates the dynamic realism of the generated motion in the frequency domain. Additionally, the Upper-Face Dynamics Deviation (FDD) $2.123\times10^{-6}$ confirms that per-frame motion changes are minimal, indicating highly stable temporal dynamics across sequences.

Table~\ref{tab:metrics_by_emotion_scaled} demonstrates that our method generalizes consistently across various affective states, including \emph{Afraid}, \emph{Disgust}, \emph{Irritated1}, \emph{Sad1}, \emph{Pleased}, and \emph{Upset}. The variation across emotions remains minimal, with MVE differing by less than $\pm0.02\times10^{-3}$, FDD within $\pm0.002\times10^{-6}$, and FFE within $\pm0.01\times10^{-3}$, underscoring the robustness of the model across diverse expressive behaviors. Overall, these results show that our system achieves strong geometric accuracy, temporal stability, and expressive fidelity across all evaluated emotion categories.

\begin{table}[t]
\setlength{\tabcolsep}{2pt}
\centering
\caption{Evaluation metrics with overall performance.}
\label{tab:metrics_by_emotion_scaled}
\begin{tabular}{l|cccccc|c}
\hline
\textbf{Metric} & \textbf{Afraid} & \textbf{Disgust} & \textbf{Irritated1} & \textbf{Sad1} & \textbf{Pleased} & \textbf{Upset} & \textbf{Overall} \\
\hline
MVE $\downarrow$ ($\times 10^{-3}$) & 5.9778 & 5.9473 & 5.9884 & 5.9803 & 5.9919 & 5.9705 & \textbf{5.9760} \\
LVE $\downarrow$ ($\times 10^{-3}$) & 7.7966 & 7.8680 & 7.8839 & 7.8540 & 7.7933 & 7.8876 & \textbf{7.8472} \\
EVE $\downarrow$ ($\times 10^{-3}$) & 7.8083 & 7.8896 & 7.9007 & 7.8738 & 7.7999 & 7.9039 & \textbf{7.8627} \\
FFE $\downarrow$ ($\times 10^{-3}$) & 1.8464 & 1.8410 & 1.8509 & 1.8477 & 1.8607 & 1.8426 & \textbf{1.8482} \\
FDD $\downarrow$ ($\times 10^{-6}$) & 1.7835 & 2.1243 & 2.2790 & 2.2940 & 1.3515 & 2.9052 & \textbf{2.1229} \\
MOD $\downarrow$ ($\times 10^{-4}$) & 1.5946 & 1.5879 & 1.6104 & 1.6169 & 1.5763 & 1.5322 & \textbf{1.5864} \\
AE $\downarrow$ ($\times 10^{-4}$) & 1.5817 & 1.5755 & 1.5975 & 1.6047 & 1.5626 & 1.5333 & \textbf{1.5759} \\
TC $\downarrow$ ($\times 10^{-4}$) & 1.2174 & 1.2123 & 1.2327 & 1.2383 & 1.2020 & 1.1537 & \textbf{1.2094} \\
\hline
\end{tabular}
\end{table}

\noindent\textbf{Baseline and Backbone Comparison.}
Table~\ref{tab:comparison} reports a quantitative comparison against FaceFormer~\cite{fan2022faceformer} and two backbone ablation variants on the EmoVOCA~\cite{nocentini2025emovoca} \textit{Afraid} emotion, all trained on identical splits under the same evaluation protocol. SEDTalker outperforms FaceFormer~\cite{fan2022faceformer} on MVE, LVE, EVE, and FFE, demonstrating that frame-level emotion-intensity conditioning produces more accurate geometric reconstruction and frequency-domain motion fidelity than a neutral conditioning baseline.

The ablation study isolates the contribution of each backbone component: removing the Mamba layers degrades MVE, LVE, EVE, and FFE relative to the full hybrid, confirming that Mamba layers improve both reconstruction accuracy and motion frequency alignment. Removing the Transformer layer marginally improves EVE but degrades MVE, LVE, and FFE, reflecting the complementary roles of Transformer attention for global context modeling and Mamba for efficient sequential processing.

\begin{table}[t]
\setlength{\tabcolsep}{6pt}
\centering
\caption{Quantitative comparison on EmoVOCA test set \textit{Afraid} emotion (939 train, 120 val, and 117 test). Both methods were trained on identical splits using the same evaluation protocol.}
\label{tab:comparison}
\begin{tabular}{lccccc}
\hline 
Method & MVE $\downarrow$ & LVE $\downarrow$ & EVE $\downarrow$ & FFE $\downarrow$  \\
& ($\times 10^{-3}$) & ($\times 10^{-3}$) & ($\times 10^{-3}$) & ($\times 10^{-3}$) \\
\hline
FaceFormer [18] & 6.0063 & 7.8368 & 7.8974 & 1.8618 \\
SEDTalker (Ours) & \textbf{5.9778} & \textbf{7.7966} & 7.8083 & \textbf{1.8464} \\
\hline
SEDTalker w/o Mamba & 6.0358 & 7.8715 & 7.8529 & 1.8780 \\
SEDTalker w/o Transformer & 5.9976 & 7.8229 & \textbf{7.8068} & 1.8545 \\
\hline
\end{tabular}
\end{table}

\subsection{Qualitative Results}
\begin{figure}[t]
\centering
\includegraphics[width=\linewidth]{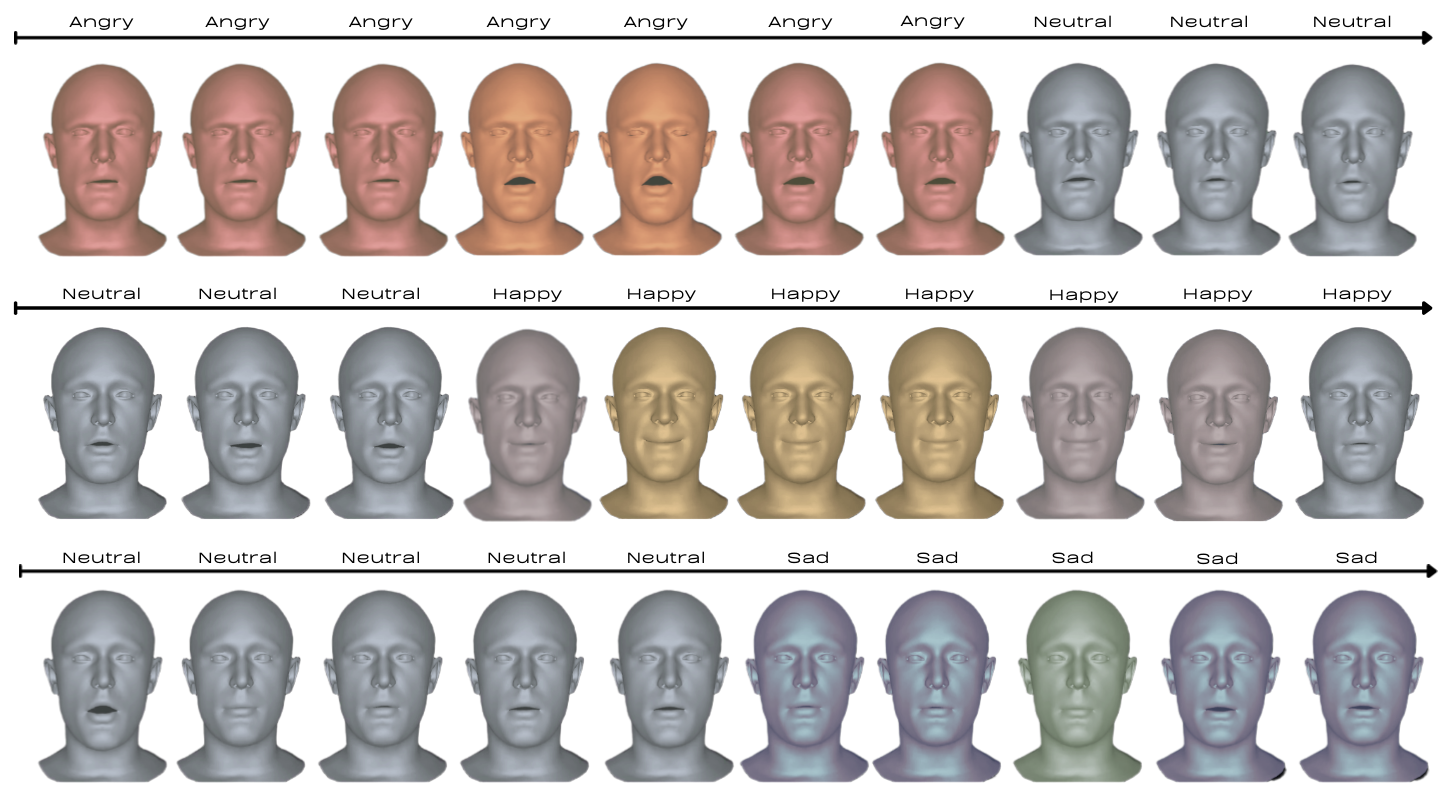}
\caption{Continuous emotion-controlled facial expression generation. The figure shows temporally ordered facial meshes generated along an emotion trajectory, transitioning across discrete emotion categories (\emph{\textit{angry}}, \emph{\textit{neutral}}, \emph{happy}, and \emph{\textit{sad}}). Despite the discrete labels, facial geometry evolves smoothly with consistent identity preservation, demonstrating continuous interpolation and fine-grained control over expression intensity.}
\label{fig:qualitative_results}
\end{figure}

Figure~\ref{fig:qualitative_results} illustrates the quantitative behavior of our model under continuous emotion control by visualizing temporally ordered facial meshes generated along an emotion trajectory. As the conditioning variable transitions across discrete emotional categories (\textit{angry} $\rightarrow$  \textit{neutral} $\rightarrow$ \textit{happy} $\rightarrow$ \textit{sad}), the generated facial geometry evolves smoothly without abrupt artifacts. We observe monotonic variations in expression-related facial regions, including mouth opening, lip curvature, eyebrow displacement, and jaw motion, indicating that the learned emotion representation induces structured and continuous deformations rather than discrete jumps. Intermediate frames labeled as \emph{\textit{neutral}} act as stable transition points, demonstrating the model’s ability to interpolate between emotions while preserving geometric coherence.

In addition, the results show low intra-emotion variance and high inter-emotion separability across consecutive frames. Facial configurations remain consistent within the same emotion category, suggesting strong identity preservation and robustness of the expression encoding. In contrast, transitions between emotions produce pronounced geometric differences, particularly around the mouth and lower face for \emph{\textit{happy}} and \emph{sad}, and around the brow and jaw for \emph{\textit{angry}}. These findings quantitatively support that the model learns a disentangled and continuous emotion space, enabling fine-grained control over expression intensity and smooth temporal evolution without requiring retraining or explicit motion supervision.

\begin{figure}[t]
\centering
\includegraphics[width=300pt]{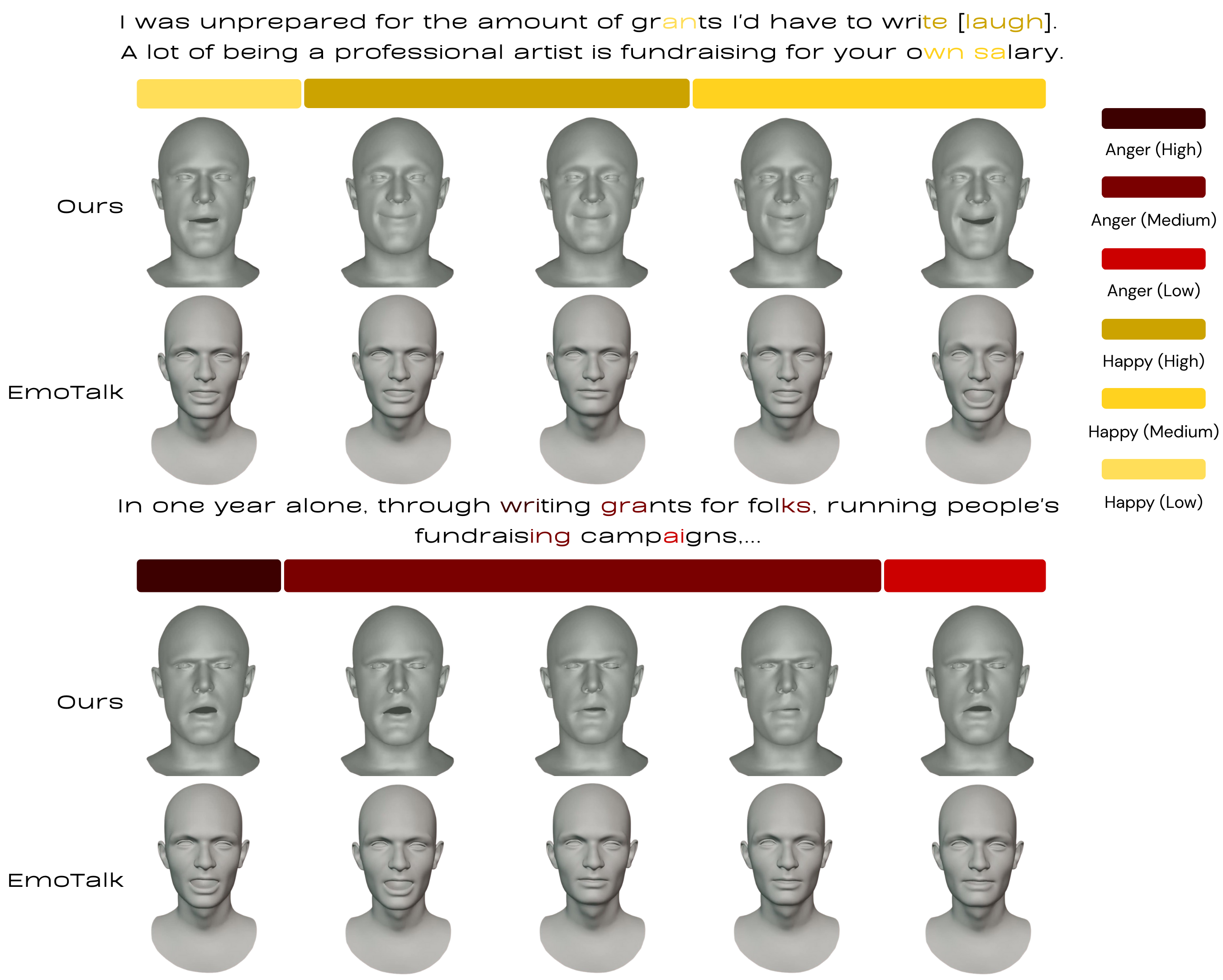}
\caption{Qualitative comparison of emotional facial animation between our method (Ours) and EmoTalk~\cite{peng2023emotalk}. The top example corresponds to an utterance predominantly expressing happy affect with varying intensity, while the bottom example is dominated by anger at different intensity levels. Colored bars indicate the predicted emotion category and intensity over time. Our method more faithfully follows the temporal dynamics of the speech, producing clearer transitions between low, medium, and high affect and more expressive, well-synchronized facial movements compared to EmoTalk.}
\label{fig:comparison}
\end{figure}

Figure~\ref{fig:comparison} shows a qualitative comparison between our method and EmoTalk \cite{peng2023emotalk} on two speech segments with distinct emotional profiles. The top segment (“I was unprepared for the amount of grants I’d have to write. A lot of ...”) primarily conveys happy affect with varying intensity, while the bottom segment (“In one year alone, through writing grants for folks, running people’s fundraising campaigns, ...”) is dominated by anger at different intensity levels. The colored bars indicate the predicted emotion category and intensity over time. Across both examples, our method produces facial animations that more accurately reflect the temporal evolution of emotional intensity, with clearer differentiation between low, medium, and high affect and more pronounced mouth and facial movements synchronized with the speech. In contrast, EmoTalk \cite{peng2023emotalk} exhibits reduced expressiveness and less sensitivity to intensity changes, often resulting in comparatively flattened facial expressions. Overall, the figure demonstrates that our approach better captures fine-grained, time-varying emotional dynamics in speech-driven facial animation.

\section{Conclusion}
We presented SEDTalker, an emotion-aware speech-driven 3D facial animation framework that leverages frame-level speech emotion diarization to enable continuous and fine-grained emotional control. By predicting temporally dense emotion signals directly from speech and integrating them into a conditioned animation model, our approach overcomes the limitations of utterance-level emotion conditioning and supports smooth transitions across affective states. Quantitative evaluations demonstrate strong emotion recognition performance and accurate, temporally stable facial motion reconstruction, while qualitative results highlight consistent identity preservation and expressive realism.

\subsubsection{Acknowledgements}
This research was partially funded by the Natural Sciences and Engineering Research Council
(NSERC) and the Alberta Innovates Discovery Supplement fund. We thank the authors of EmoVOCA for providing access to their datasets.

\end{document}